\documentclass[letterpaper, 10 pt, journal, twoside]{IEEEtran}

\usepackage[pdftex]{graphicx}
\usepackage{amsmath}
\usepackage{cite}
\usepackage{booktabs}

\usepackage{float}
\usepackage{color}
\usepackage[table]{xcolor}
\usepackage{colortbl}
\usepackage{hyperref}
\usepackage{booktabs}
\usepackage{multirow}
\usepackage{bbding}

\begin{document}

\title{MOSE: Monocular Semantic Reconstruction Using \\ NeRF-Lifted Noisy Priors}

\author{Zhenhua Du$^{1}$, Binbin Xu$^{2}$, Haoyu Zhang$^{1}$, Kai Huo$^{1}$, Shuaifeng Zhi$^{1, \dagger}$% <-this % stops a space
\thanks{Manuscript received: April, 7, 2024; Revised July, 10, 2024; Accepted August, 23, 2024.}%Use only for final RAL version
\thanks{This paper was recommended for publication by Editor Cesar Cadena Lerma upon evaluation of the Associate Editor and Reviewers' comments.}
\thanks{Research presented in this paper has been partly supported by the NSFC (No. 62201603), the CPSF (No. 2023TQ0088, No. GZC20233539), and the Research Program of NUDT (No. ZK22-04).} %Use only for final RAL version
\thanks{$^{1}$Zhenhua Du, Haoyu Zhang, Kai Huo, Shuaifeng Zhi$^{\dagger}$ (\textit{corresponding author}) are with National University of Defense Technology. \newline
Emails: {\tt\footnotesize dzh11@nudt.edu.cn, zhanghaoyu19@nudt.edu.cn, huokai2001@163.com, zhishuaifeng@outlook.com.}}%
\thanks{$^{2}$Binbin Xu: {\tt\footnotesize xu.binbin@outlook.com.}}%
\thanks{Source code: \href{https://github.com/ZhenhuaDu11/Mose}{https://github.com/ZhenhuaDu11/Mose}.}
\thanks{Digital Object Identifier (DOI): see top of this page.}
}

\markboth{IEEE Robotics and Automation Letters. Preprint Version. Accepted September, 2024}
{Du \MakeLowercase{\textit{et al.}}: MOSE: Monocular Semantic Reconstruction Using NeRF-Lifted Noisy Priors} 

\maketitle

%%%%%%%%%%%%%%%%%%%%%%%%%%%%%%%%%%%%%%%%%%%%%%%%%%%%%%%%%%%%%%%%%%%%%%%%%%%%%%%%
\begin{abstract}
Accurately reconstructing dense and semantically annotated 3D meshes from monocular images remains a challenging task due to the lack of geometry guidance and imperfect view-dependent 2D priors. Though we have witnessed recent advancements in implicit neural scene representations enabling precise 2D rendering simply from multi-view images, there have been few works addressing 3D scene understanding with monocular priors alone. In this paper, we propose MOSE, a neural field semantic reconstruction approach to lift inferred image-level noisy priors to 3D, producing accurate semantics and geometry in both 3D and 2D space. The key motivation for our method is to leverage generic class-agnostic segment masks as guidance to promote local consistency of rendered semantics during training. With the help of semantics, we further apply a smoothness regularization to texture-less regions for better geometric quality, thus achieving mutual benefits of geometry and semantics. Experiments on the ScanNet dataset show that our MOSE outperforms relevant baselines across all metrics on tasks of 3D semantic segmentation, 2D semantic segmentation and 3D surface reconstruction. 
\end{abstract}


%%%%%%%%%%%%%%%%%%%%%%%%%%%%%%%%%%%%%%%%%%%%%%%%%%%%%%%%%%%%%%%%%%%%%%%%%%%%%%%%
\section{INTRODUCTION}
\IEEEPARstart{C}{omprehensively} understanding the high-level semantics of 3D scenes is crucial for various downstream applications, including augmented reality, robot navigation, and autonomous driving. Given the prohibitive costs of acquiring dense 3D scans and corresponding annotations, it would always be desirable if a dense semantic 3D map could be achieved simply from its multi-view 2D observations, benefiting from the widely available training corpora. There have been great related advancements using SFM and RGB-D vSLAM systems, projecting 2D semantic predictions into 3D space and fusing them via prescribed rules like Bayesian fusion \cite{McCormac:etal:ICRA2017, Narita:etal:IROS2019}. Nevertheless, addressing this task with monocular cues remains challenging, especially when confronted with inconsistent 2D semantic labels and inaccurately reconstructed geometry. As an attempt in addressing this task, in this paper, our objective is to reconstruct a consistent 3D semantic map of indoor scenes only with imperfect multi-view 2D priors. 

Recently proposed neural implicit representations (NeRF) \cite{Mildenhall:etal:ECCV2020} have shown impressive performance in capturing intricate appearance and geometric details from only RGB images, with a clear 3D awareness of multi-view consistency. Semantic-NeRF \cite{Zhi:etal:ICCV2021} employs an extra MLP to represent the semantic field of the scene, demonstrating how 2D semantic predictions benefit from the self-similarity inherent in the compact scene encodings. However, due to the lack of sufficient surface constraints, Semantic-NeRF and its variants \cite{siddiqui:CVPR2023:panopticlifting,Bhalgat:2023:contrastivelift} often produce floaters when extracting 3D semantic maps. VolSDF \cite{yariv:2021:volsdf} and NeuS \cite{wangNeuS2021}, on the other hand, improve geometry quality of NeRF by parameterizing the density filed as an SDF field, and alternatively use depth and normal priors for even higher quality \cite{wang2022neuris,li:ICCV2023:rico}. Built upon the work of \cite{Zhi:etal:ICCV2021} and \cite{yariv:2021:volsdf}, Manhattan-SDF \cite{Guo:etal:CVPR2022} learns a joint representation of scene geometry and semantics motivated by Manhattan-world assumption within indoor scenes. Though improved monocular 3D semantic mapping is achieved, \cite{Guo:etal:CVPR2022} only considers three coarse semantic classes (wall, floor, and others), which limits its applicability in cases demanding detailed categorization. 

\begin{figure}[!t]
    \centering
    \includegraphics[width=1\linewidth]{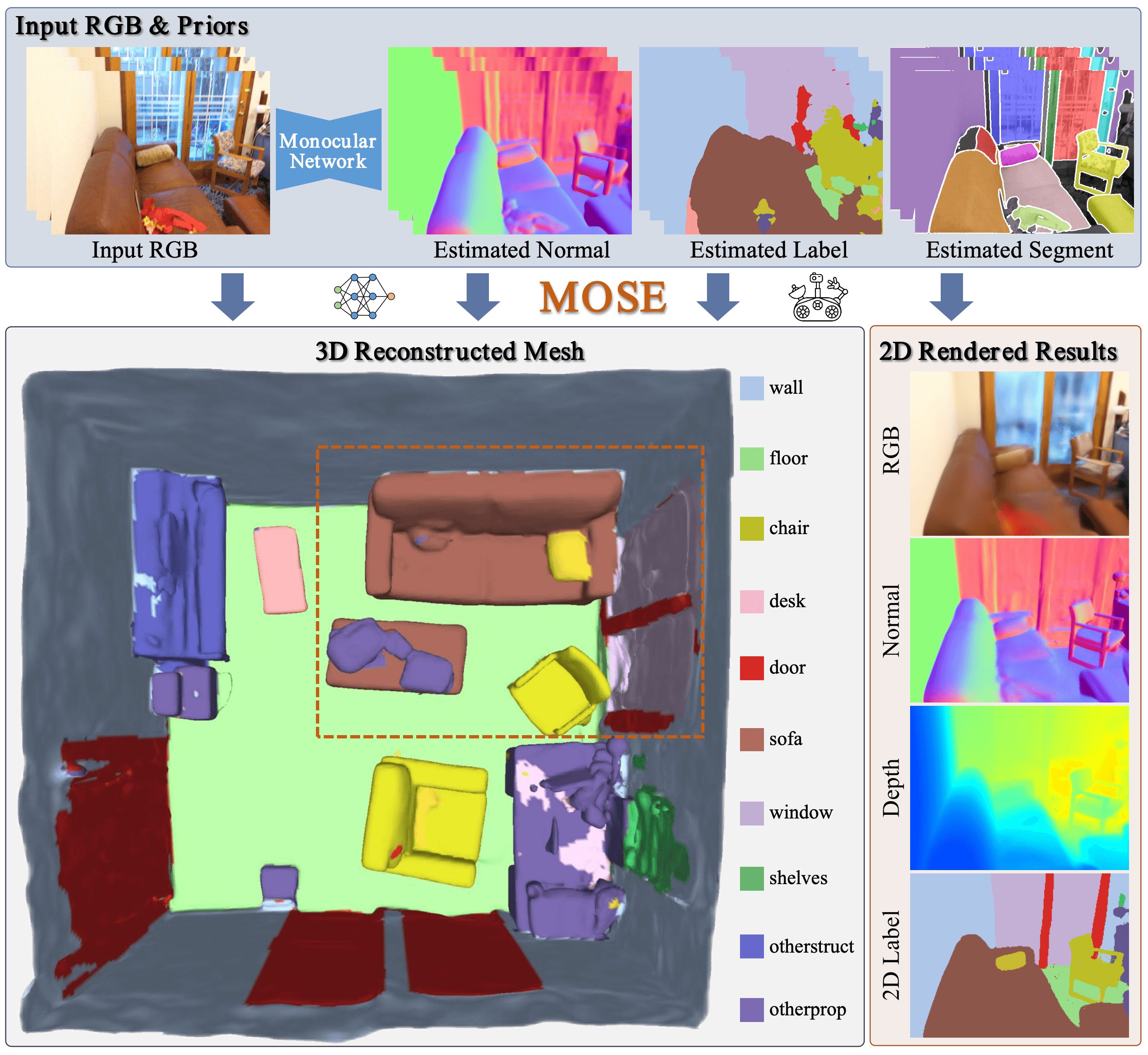}
    \vspace{-0.6cm}
    \caption{\textbf{3D indoor semantic reconstruction}. Taking RGB images and noisy 2D scene priors from monocular networks (upper portion), our method MOSE, is able to reconstruct the 3D smooth semantic map of the scene and render 2D associated results (bottom portion).}
    \label{fig:Into:overview}
    \vspace{-0.4cm}
\end{figure}

In this paper, we propose MOSE, to concurrently achieve high-quality 3D geometry scans as well as finer-grained semantic labeling purely from a series of 2D images along with imperfect priors including normals, semantics and segments. Our motivation comes from the observation that there is a lack of effective region-wise coherence within learned object semantics and texture-less geometry, leading to obvious defects in terms of incongruous semantic patches and fractured surfaces. Our MOSE exploits local segments as basis to encourage further smoothness of semantics, and furthermore relies on semantics to guide the learning process in texture-less regions. With negligible overhead during training, our approach produces accurate semantic 3D scans and surpasses existing baselines with a clear margin.

To summarize, we propose MOSE, an implicit monocular semantic reconstruction system, whose key contributions are listed as follows:
\begin{itemize}
    \item We introduce a neural semantic reconstruction system capable of reconstructing smooth 3D semantic maps only from images and noisy 2D scene priors, which obtains state-of-the-art 3D semantic understanding performance on the challenging ScanNet scenes.
    \item To better utilize inconsistent semantic supervision, a locally-consistent fusion strategy using class-agnostic image segments (e.g., SAM\cite{Kirillov:etal:SAM:ARXIV2023}, super-pixels\cite{felzenszwalb2004efficient}) is proposed, obtaining smooth and accurate semantics.
    \item To mitigate geometric degradation in texture-less regions, we introduce a semantically-weighted geometric regularization term to encourage stronger smoothness on dominant semantic structures.
\end{itemize}

%%%%%%%%%%%%%%%%%%%%%%%%%%%%%%%%%%%%%%%%%%%%%%%%%%%%%%%%%%%%%%%%%%%%%%%%%%%%%%%%
\section{RELATED WORK}
\noindent \textbf{Neural Implicit Representation.} There is a trend to encode a scene into an implicit function by training a coordinate-based neural network \cite{Mescheder:etal:CVPR2019:occupancynet,Park:etal:CVPR2019:deepsdf}. In particular, Neural Radiance Field (NeRF) \cite{Mildenhall:etal:ECCV2020} has opened up a line of research of representing scenes as volumetric density fields to learn geometry and appearance simultaneously. However, the density-based methods face challenges when extracting high-fidelity surfaces. Following-up works Neus \cite{wangNeuS2021} and VolSDF \cite{yariv:2021:volsdf} correlate the signed distance function (SDF) with density, enabling accurate surface reconstruction. To reconstruct larger scenes, estimated geometric priors such as depth and normal maps, are often introduced to aid representation learning \cite{wang2022neuris, park:ICLR2024:h2oSDF, li:ICCV2023:rico}. Among these methods, NeuRIS\cite{wang2022neuris} effectively reconstructs room-scale scenes by adaptively using noisy normal priors. However, due to the inherent shape ambiguity of implicit representation when fitting multi-view training images \cite{Zhang:etal:ARXIV2020NeRF++}, degradation still occurs in texture-less regions even when the rendering results align well with 2D supervision. In this work, we propose a compact implicit representation that captures scene appearance, geometry and semantics from only RGB and 2D noisy neural priors. 

\noindent \textbf{Neural Semantic Reconstruction.} Most traditional semantic reconstruction methods \cite{McCormac:etal:ICRA2017, Narita:etal:IROS2019} typically project 2D predicted labels into a fused semantic map. Due to the lack of perceptual awareness of neighboring pixels, they often struggle to handle complex scenarios. Utilizing neural networks \cite{He:CVPR2016:resnet}, SceneCode \cite{Zhi:etal:CVPR2019} encodes the scene into compact latent codes and jointly optimize the geometry and semantics. Inspired by the success of NeRF \cite{Mildenhall:etal:ECCV2020}, Semantic-NeRF \cite{Zhi:etal:ICCV2021} and its following-up works \cite{siddiqui:CVPR2023:panopticlifting, Bhalgat:2023:contrastivelift} attempt to learn a 3D consistent semantic field via lifting 2D semantic labels. However, these methods solely learn the density field and face challenges in extracting a high-quality 3D semantic map directly. Adopting the surface representation of VolSDF \cite{yariv:2021:volsdf}, Manhattan-SDF \cite{Guo:etal:CVPR2022} simultaneously learns scene geometry and semantics, enabling monocular 3D semantic mapping of indoor scenes. However, \cite{Guo:etal:CVPR2022} is limited to three coarse semantic classes and exhibits fluctuating surface geometry. In this work, we propose MOSE, aiming to simultaneously achieve accurate 3D geometry scans as well as smooth semantic labeling.

%%%%%%%%%%%%%%%%%%%%%%%%%%%%%%%%%%%%%%%%%%%%%%%%%%%%%%%%%%%%%%%%%%%%%%%%%%%%%%%%
\section{METHOD}

\begin{figure*}[!t]
    \centering
    \includegraphics[width=1\linewidth]{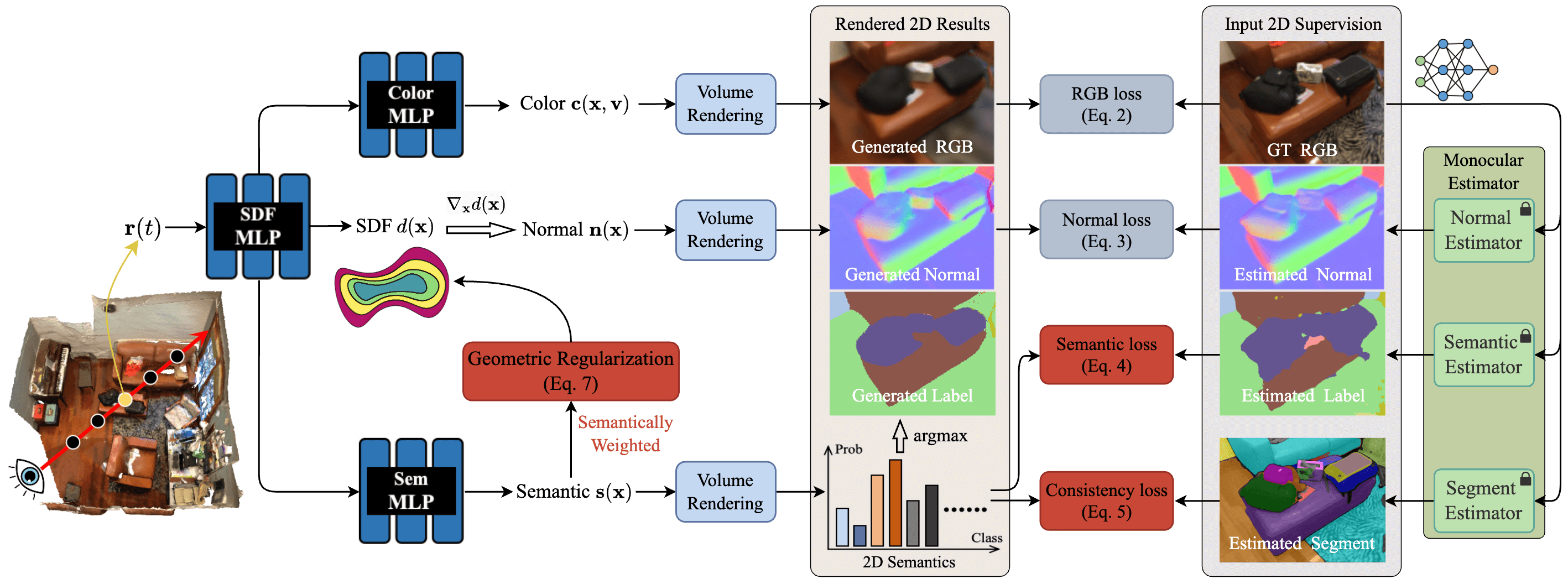}
    \vspace{-0.8cm}    \caption{\textbf{Overview of MOSE.} Utilizing RGB images and estimated normals, semantic labels, as well as segment masks, MOSE learns the color field, signed distance function (SDF) field and semantic field of the scene through an implicit neural representation. To address the discontinuity of 2D semantic predictions, we propose a locally-consistent fusion strategy (Sec. \ref{section 3.2}) leveraging 2D segmentation techniques. Semantically-weighted geometric regularization (Sec. \ref{section 3.3}) is further introduced to bring benefits to both the SDF field and semantic field.}
    \label{fig:Method:pipeline}
    \vspace{-0.4cm}
\end{figure*}

Scanning the scene with a monocular camera, our objective is to reconstruct an accurate and smooth 3D semantic map of indoor scenes. As illustrated in Fig. \ref{fig:Into:overview}, MOSE takes as input posed RGB images $\{\mathcal{I}_k\}$, as well as noisy 2D scene priors from off-the-shelf predictors, including semantics class labels $\{\mathcal{S}_k\}$, estimated normals $\{\mathcal{N}_k\}$ and class-agnostic segment masks $\{\mathcal{M}_k\}$. To faithfully reconstruct semantic 3D scans, we encode scene geometry, appearance and semantics into a neural field representation and conduct joint optimization to achieve mutual benefits (Sec. \ref{section 3.1}). To overcome the inherent discontinuity of 2D semantic predictions \cite{chen2018deeplabV3+, cheng2022mask2former}, we propose a locally-consistent fusion strategy (LCF) (Sec. \ref{section 3.2}) to improve semantics' local coherence and smoothness by leveraging generic 2D segment masks as a prior, which has been found to be particularly useful in producing accurate object-level semantics. In addition, we show further benefits of acquiring accurate semantics by improving geometry quality using the proposed semantically-weighted geometric regularization (SGR) (Sec. \ref{section 3.3}).
% object-similarity , leveraging 2D segment model priors. 

%%%%%%%%%%%%%%%%%%%%%%%%%%%%%%%%%%%%%%%%%%%
\vspace{-0.2cm}
\subsection{Scene Representation and Rendering} \label{section 3.1}

MOSE adopts a NeRF-based representation to compactly encode scene appearance, geometry and semantics (see Fig. \ref{fig:Method:pipeline}). Specifically, three MLPs are used: for spatial points $\mathbf{x}=(x, y, z)$ and their viewing direction $\mathbf{v} = (\theta, \phi)$, a color MLP encodes the appearance as a color field $\mathbf{c}(\mathbf{x}, \mathbf{v})$, an SDF MLP represents the geometry as a signed distance function (SDF) field $d(\mathbf{x})$, and a semantic MLP encodes its semantic class label as a semantic field $\mathbf{s}(\mathbf{x})$.

To learn SDF field, we adopt the neural surface representation from NeuS \cite{wangNeuS2021}, which transforms the volume density $\sigma(\mathbf{x})$ to an SDF value $d(\mathbf{x})$. We sample 3D points $\mathbf{x}_i$ along the camera ray $\mathbf{r}$, and utilize volume rendering to accumulate its color $\mathbf{c}_i$, normal $\mathbf{n}_i$ and semantic logits $\mathbf{s}_i$:
\begin{equation}
    \hat{\mathbf{C}}(\mathbf{r}) = \sum_{i=1}^N w_i \mathbf{c}_i,
    \hat{\mathbf{N}}(\mathbf{r}) = \sum_{i=1}^N w_i \mathbf{n}_i,
    \hat{\mathbf{S}}(\mathbf{r}) = \sum_{i=1}^N w_i \mathbf{s}_i,
    \vspace{-0.1cm}
\end{equation}
where $\hat{\mathbf{C}}$, $\hat{\mathbf{N}}$, $\hat{\mathbf{S}}$ are the predicted 2D color, normal and semantic logits of ray $\mathbf{r}$, respectively; $\alpha_i =1-\text{exp}(-\sigma_i \delta_i)$ is the discrete opacity of $i$-th sampled point, $w_i = \alpha_i \prod_{j=1}^{i-1} (1- \alpha_j)$ represent its transmittance, $\delta_i = \Vert x_{i+1}-x_{i} \Vert_2$ is the distance between adjacent sampled points. 

We use photometric loss $\mathcal{L}_{c}$, normal loss $\mathcal{L}_n$ and semantic loss $\mathcal{L}_s$ to supervise the network: 
\begin{equation}\label{RGB loss}
    \mathcal{L}_{c} = \sum_{\mathbf{r} \in \mathcal{R}} \Vert \hat{\mathbf{C}}(\mathbf{r})- \mathbf{C}(\mathbf{r}) \Vert_1,
\end{equation}
\vspace{-0.1cm}
\begin{equation}\label{normal loss}
    \mathcal{L}_{n} = \sum_{\mathbf{r} \in \mathcal{R}} \Vert \hat{\mathbf{N}}(\mathbf{r}) - {\mathbf{N}}(\mathbf{r}) \Vert_1 \cdot \Omega_n(\mathbf{r}), 
\end{equation}
\vspace{-0.3cm}
\begin{equation}\label{semantic loss}
    \mathcal{L}_{s} = -\sum_{\mathbf{r} \in \mathcal{R}} [\sum^L_{l=1} p^l(\mathbf{r}) \text{log} \hat{p}^l(\mathbf{r})], 
\end{equation}
where $\mathcal{R}$ are the sampled rays, $\mathbf{C}$ and $\mathbf{N}$ are the corresponding RGB color and normal prior, respectively. Following \cite{wang2022neuris}, we apply normal supervision only to pixels satisfying a pre-defined patch-wise similarity metric, indicated by $\Omega_n(\cdot)$. With a set of predefined $L$ semantic classes, $\hat{p}^l(\mathbf{r})$ is the rendered semantic probability at class $l$ after softmax normalization, and $p^l(\mathbf{r})$ is the input 2D semantic supervision.

It is worth to note that we prevent gradients from semantics (Eq. \ref{semantic loss}) to SDF MLP. We empirically observe that the volume density (i.e., SDF value) often distorts to accommodate input noisy labels, leading to significant geometric degradation. Similar finding has also been found in \cite{siddiqui:CVPR2023:panopticlifting}. As shown in Tab. \ref{Table:Ablation}, adopting this strategy benefits both surface reconstruction and semantic understanding.

%%%%%%%%%%%%%%%%%%%%%%%%%%%%%%%%%%%%%%%%%%%
\vspace{-0.2cm}
\subsection{Locally-Consistent Fusion Strategy} \label{section 3.2}

\begin{figure}[!t]
    \centering
    \includegraphics[width=1\linewidth]{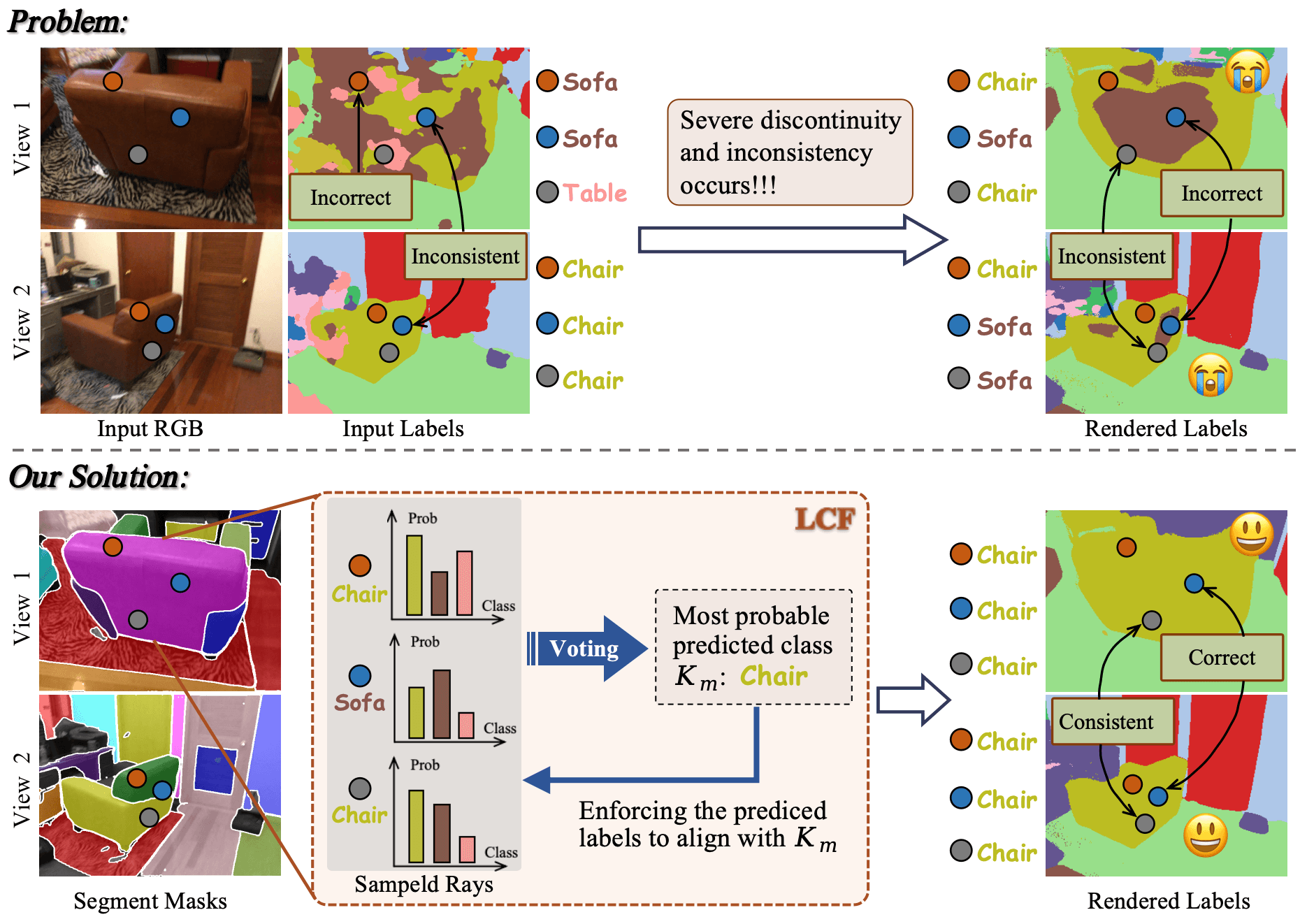}
    \vspace{-0.6cm}
    \caption{\textbf{Overview of locally-consistent fusion strategy.} Severe discontinuity and inconsistency of semantics can be observed when directly inputting noisy multi-view labels into a NeRF-based fusion system (upper part). Our LCF strategy utilizes 2D segment priors to enforce consistent and accurate semantic distributions within each segment mask (bottom part).}
    \label{fig:Method:LCF}
    \vspace{-0.4cm}
\end{figure}

Supervised by imperfect semantic predictions \cite{chen2018deeplabV3+, cheng2022mask2former}, reasonable 2D labels can be directly rendered from a joint compact scene representation \cite{Zhi:etal:ICCV2021, wang2022neuris}. However, we observe frequent discontinuity and inconsistency among semantic rendering, especially on objects shown as noisy patches and bleeding edges in Fig. \ref{fig:Method:LCF}. We suspect that the element-wise fusion lacks an explicit awareness of the local coherence of semantics. To address this issue, we propose to rely on generic images segments obeying image structures like boundaries. We introduce a locally-consistent fusion strategy (LCF), leveraging on 2D segmentation techniques (e.g. SAM\cite{Kirillov:etal:SAM:ARXIV2023}, super-pixel segmentation \cite{felzenszwalb2004efficient}), to perceive neighboring pixels and output locally consistent semantics.

Without loss of generality, we use SAM \cite{Kirillov:etal:SAM:ARXIV2023} to process each multi-view image and generate class-agnostic masks $\{\mathcal{M}_k\}$. We also ablate the choice of using super-pixels \cite{felzenszwalb2004efficient} in Sec. \ref{discussion}. During training, we encourage learned semantic distribution to be consistent within each local segment mask (Fig. \ref{fig:Method:LCF}). Specifically, sampled rays located within the same mask $m$ are clustered into a group $\mathcal{R}_m$. For each group, we vote the most probably predicted class $K_m$ by counting the number of occurrences of each class. We also tried other voting strategies like soft voting but find negligible difference. We hence stick to current simple yet efficient choice. During training, after obtaining the voted labels of $\mathcal{R}_m$, the semantic consistency loss is defined as follows: 
\begin{equation}
    \mathcal{L}_{con}=-\sum_{m=1}^M \sum_{\mathbf{r} \in \mathcal{R}_m}  \text{log} \hat{p}^{K_m}(\mathbf{r}), 
\end{equation}
where $\hat{p}^{K_m}(\mathbf{r})$ is the rendered semantic probability of the voted class $K_m$. Although 2D masks $\{\mathcal{M}_k\}$ vary across frames, we incorporate LCF in a frame-wise manner and learn a more consistent and accurate semantic field.

\subsection{Semantically-Weighted Geometric Regularization} \label{section 3.3}

\begin{figure}[!t]
    \centering
    \includegraphics[width=1\linewidth]{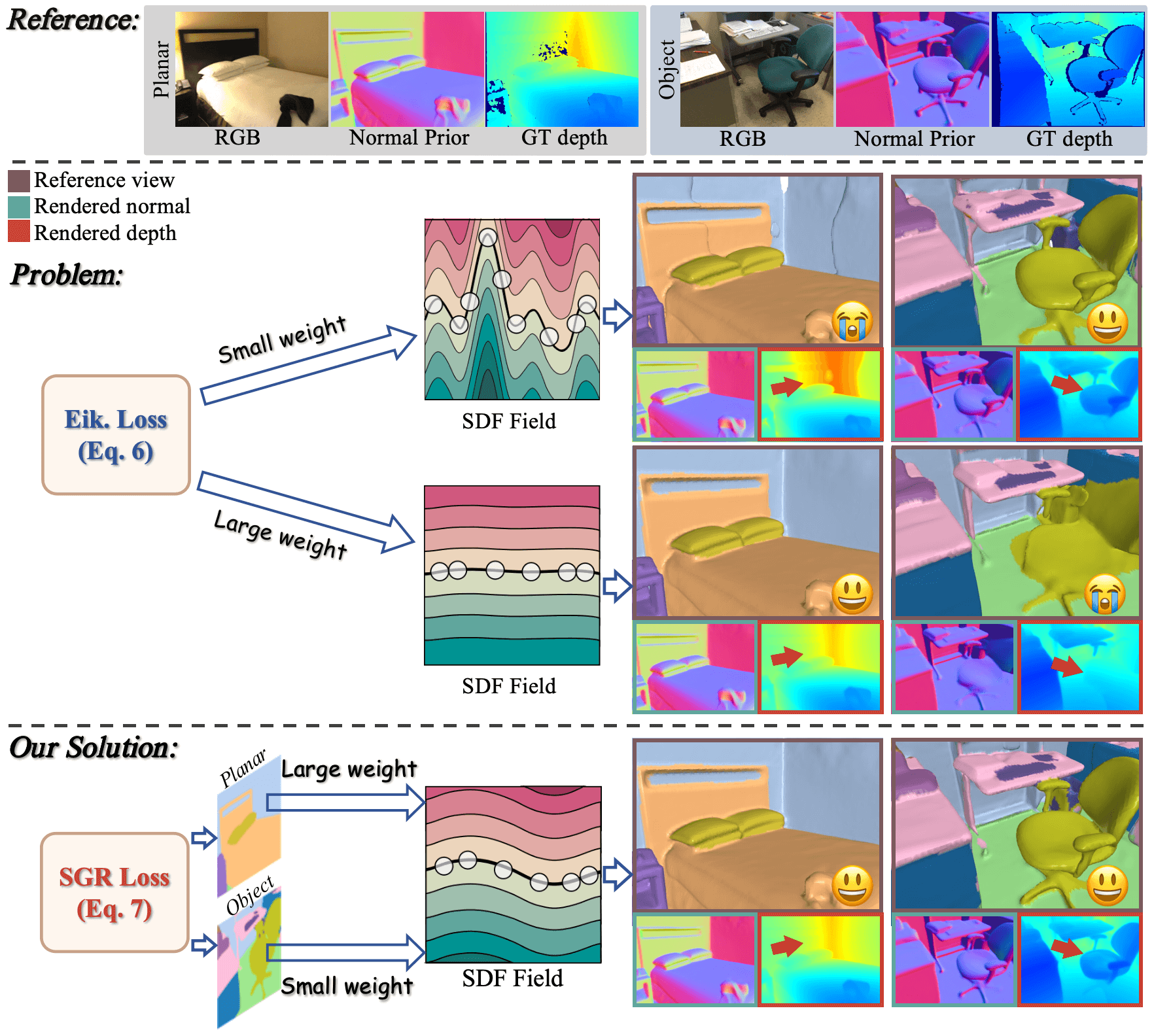}
    \vspace{-0.6cm}
    \caption{\textbf{Overview of semantically-weighted geometric regularization.} Achieving a balance between planar and object regions is challenging with the widely-used Eikonal loss \cite{gropp2020igr}: large loss weights lead to discontinuity in texture regions, while small loss weights result in loss of object details. Our proposed semantically-weighted geometric regularization (SGR) dynamically adjusts the regularization strength across different semantic classes, resulting in more accurate surface reconstruction. Semantics also benefit from the more accurate radiance field.}
    \label{fig:Method:SGR}
    \vspace{-0.4cm}
\end{figure}

Accurately reconstructing scene geometry from images alone is difficult even with normal prior, since degradation usually occurs in low-texture regions due to their inherent shape ambiguity \cite{Zhang:etal:ARXIV2020NeRF++}. As shown in Fig. \ref{fig:Method:SGR}, though the rendered normal maps have closely matched input normal priors, the reconstructed walls still exhibit obvious fragmentation. 

Recall that previous methods \cite{yariv:2021:volsdf, wangNeuS2021} usually use Eikonal loss \cite{gropp2020igr} to regularize the SDF field:
\begin{equation} \label{eikonal loss}
    \mathcal{L}_{eik} = \sum_{\mathbf{x} \in \mathcal{{X}}} (\Vert \nabla_{\mathbf{x}} d(\mathbf{x}) \Vert_2 - 1)^2,
\end{equation}
where $\mathcal{X}$ denotes a set of sampled points $\mathbf{x}$ and $d(\mathbf{x})$ is the associated SDF field. One straightforward solution to enhance the continuity of such texture-less planar regions is to increase the weighting of Eikonal loss. However, aggressively applying excessive strength to the Eikonal term worsens geometry by over-smoothing objects (Fig. \ref{fig:Method:SGR}). Thanks to the accurate semantics encouraged by LCF, we could adaptively adjust the strength of Eikonal loss based on learned semantic classes, and propose a semantically-weighted geometric regularization (SGR), defined as follows:
\begin{equation}
    \mathcal{L}_{sgr} = \sum_{\mathbf{x} \in \mathcal{{X}}} (1+\Phi(\hat{p}_{\mathbf{x}})) (\Vert \nabla_{\mathbf{x}} d(\mathbf{x}) \Vert_2 - 1)^2,
\end{equation}
where $\hat{p}_{\mathbf{x}}$ denotes the rendered multi-class semantic probability of sampled points $\mathbf{x}$, $\Phi(\hat{p}_{\mathbf{x}})$ is a semantically-adjusted weighting function, defined as:
\begin{equation}
\Phi(\hat{p}_{\mathbf{x}}) = \begin{cases}
\sum_{l \in \mathcal{P}} \hat{p}^l(\mathbf{x}) & \text{if}\ \hat{l}_{\mathbf{x}} \in \mathcal{P} \\
0 & \text{if}\  \hat{l}_{\mathbf{x}} \in \mathcal{O},
\end{cases}
\end{equation}
where $\hat{l}_{\mathbf{x}}$ is the predicted semantic label of sampled points $\mathbf{x}$, we consider $\mathcal{P}$ to be dominant indoor structures like walls, floors and ceilings, which are also planar regions, and $\mathcal{O}$ to be other object classes. We dynamically adjust the strength of our geometric regularization across different semantic classes, ensuring a stronger SDF smoothness on planar regions while preserving objects' details. Though SGR does not explicitly optimize for semantics, we observe improvements in semantics owned to better geometry quality, as shown in Tab. \ref{Table:Ablation}.

\subsection{Training and Implementation Details} \label{section 3.4}
We train our model with the following loss function:
\begin{equation}
    \mathcal{L} = w_{c} \mathcal{L}_{c} + 
                w_{n} \mathcal{L}_{n} +
                w_{s} \mathcal{L}_{s} +
                w_{con} \mathcal{L}_{con} +
                w_{sgr} \mathcal{L}_{sgr},
\end{equation}
where we set the weighting factors as $w_{c}=w_{n}=1$, $w_{s}=w_{con}=0.5$, $w_{sgr}=0.1$, respectively. We sample 512 rays for each batch and optimize using the Adam optimizer \cite{Kingma:Adam:ICLR2015} with a learning rate of $2\times10^{-4}$. Our MLPs are trained on 1 NVIDIA V100 GPU for 160,000 iterations ($\sim$10 hours). Hyperparameters of MLPs are similar to \cite{wang2022neuris} and \cite{Guo:etal:CVPR2022} for a fair comparison. To obtain the 3D semantic map, we use the Marching Cubes algorithm \cite{Lorensen:Cline:SIGGRAPH1987:marchingcube} for extracting surface mesh and compute the semantic labels of each vertex on the mesh. 

%%%%%%%%%%%%%%%%%%%%%%%%%%%%%%%%%%%%%%%%%%%%%%%%%%%%%%%%%%%%%%%%%%%%%%%%%%%%%%%%
\section{EXPERIMENTS}
\begin{figure*}[!t]
    \centering
    \includegraphics[width=1\linewidth]{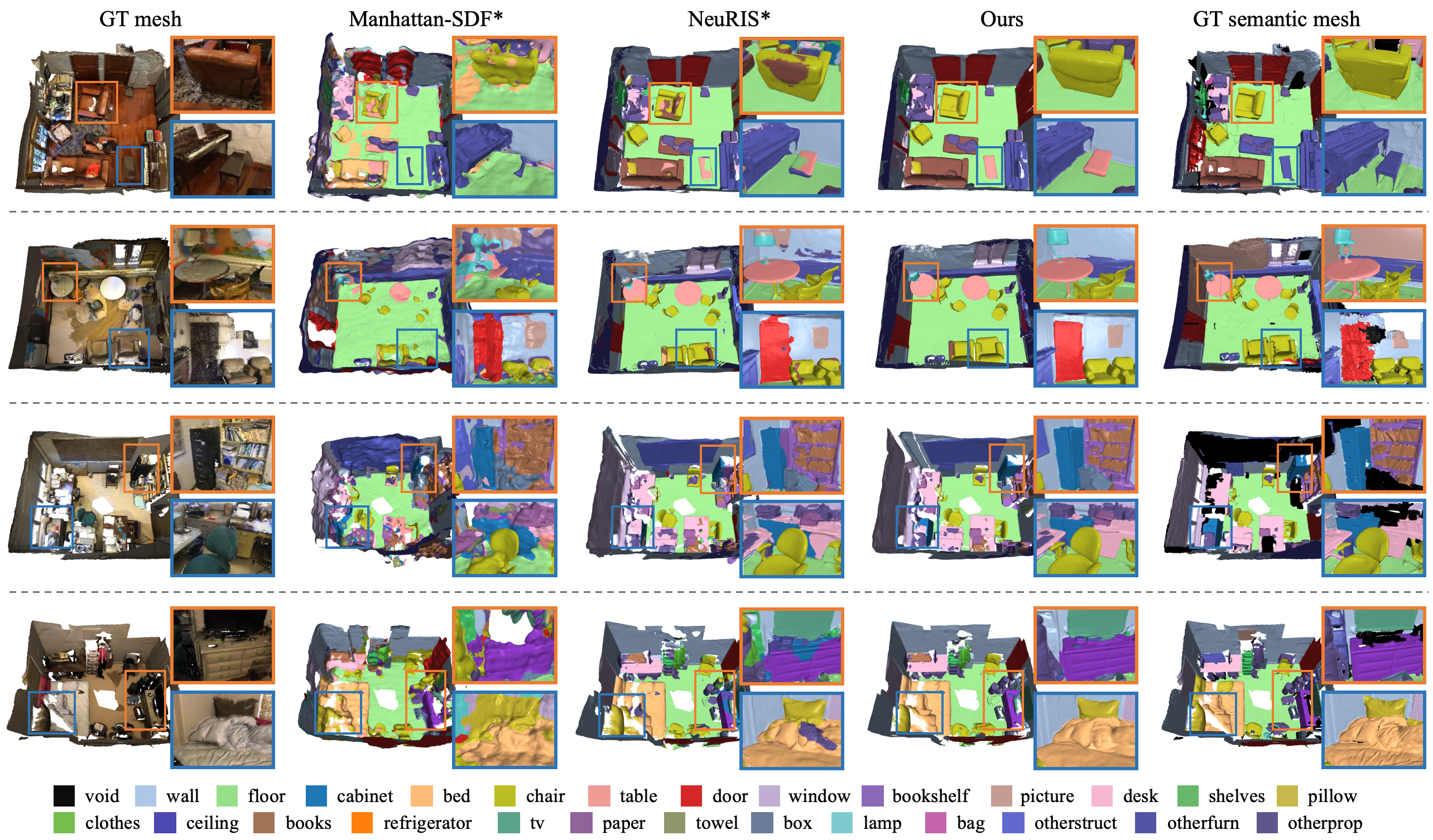}
    \vspace{-0.6cm}
    \caption{\textbf{Qualitative comparisons of 3D semantic reconstruction results.} Our method is able to produce smoother 3D semantic map and align well with GT results, while Manhattan-SDF* and NeuRIS* exhibits severe inconsistencies of semantic labels. }
    \label{fig:Comparison:3DSemseg}
\end{figure*}

\begin{figure*}[!t]
\vspace{-0.1cm}
    \centering
    \includegraphics[width=1\linewidth]{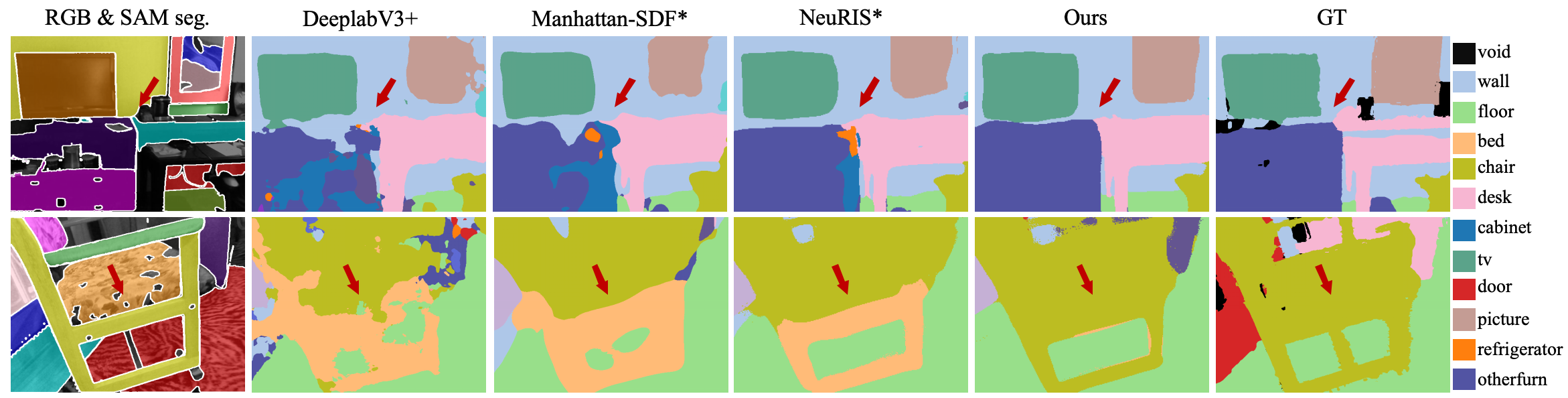}
    \vspace{-0.6cm}
    \caption{\textbf{Qualitative comparisons of 2D rendering semantic results.} Inputting noisy labels from DeeplabV3+ \cite{chen2018deeplabV3+}, both Manhattan-SDF* and NeuRIS* struggle to render reasonable 2D semantic images. Utilizing SAM \cite{Kirillov:etal:SAM:ARXIV2023} segments, our method outperforms baselines with more consistent results. }
    \label{fig:Comparison:2DSemseg}
    \vspace{-0.4cm}
\end{figure*}

\subsection{Datasets, metrics and baselines}
\noindent \textbf{Datasets.} We validate our approach on the popular ScanNet \cite{dai2017scannet} dataset. ScanNet is a large-scale real-world indoor RGB-D video dataset consisting of 1613 indoor scenes with ground-truth camera poses, surface reconstructions and semantic annotations. Following previous works \cite{Guo:etal:CVPR2022, wang2022neuris}, we select 8 representative scenes and use provided camera poses to perform experiments. As for training and evaluation of semantics, we adopt the widely used NYU-40 convention. Input images are resized to $640\times480$ pixels as well as other monocular priors, and we equally sampled 10\% images of each scene to train our model \cite{wang2022neuris}.

\noindent \textbf{Neural Priors.} For normal priors, we use the SNU network \cite{bae2021estimating} provided by NeuRIS \cite{wang2022neuris}. For input semantic labels, we use DeepLabV3+ \cite{chen2018deeplabV3+} with a ResNet-101 backbone \cite{He:CVPR2016:resnet} and re-train it on the ScanNet-frames-25k dataset (a subset of ScanNet) for 50 epoches. Both normal network and semantic networks are trained on the training split of ScanNet \cite{dai2017scannet} and evaluated on 8 scenes from the validation set. For generic 2D segment priors, we use the standard SAM \cite{Kirillov:etal:SAM:ARXIV2023} and only keep segments larger than 4000 pixels to encourage consistency in a wider locality. In Sec. \ref{discussion}, we also validate our method using Mask2Former \cite{cheng2022mask2former} and classical graph-based super-pixels \cite{felzenszwalb2004efficient} as semantics and segments predictions, respectively.

\noindent \textbf{Metrics.} Our method focuses on the task of 3D semantic reconstruction, 2D semantic segmentation and 3D geometry reconstruction from multi-view images. For 3D semantic segmentation, we transfer semantic labels from the reconstructed 3D mesh to the ground-truth mesh for the evaluation purpose. For 2D semantic segmentation, we render 2D semantic images at training viewpoints and compare them to ground-truth labels. We adopt total pixel accuracy (Acc), average pixel accuracy (mAcc), and mean class-wise intersection over union (mIoU) as the semantic evaluation metrics. For geometry reconstruction, we follow the evaluation procedure of \cite{Guo:etal:CVPR2022}, utilizing six standard metrics: accuracy (Acc), completeness (Comp), precision (Prec), recall, F-score and chamfer distance (CD).

\begin{figure*}[!t]
    \centering
    \includegraphics[width=1\linewidth]{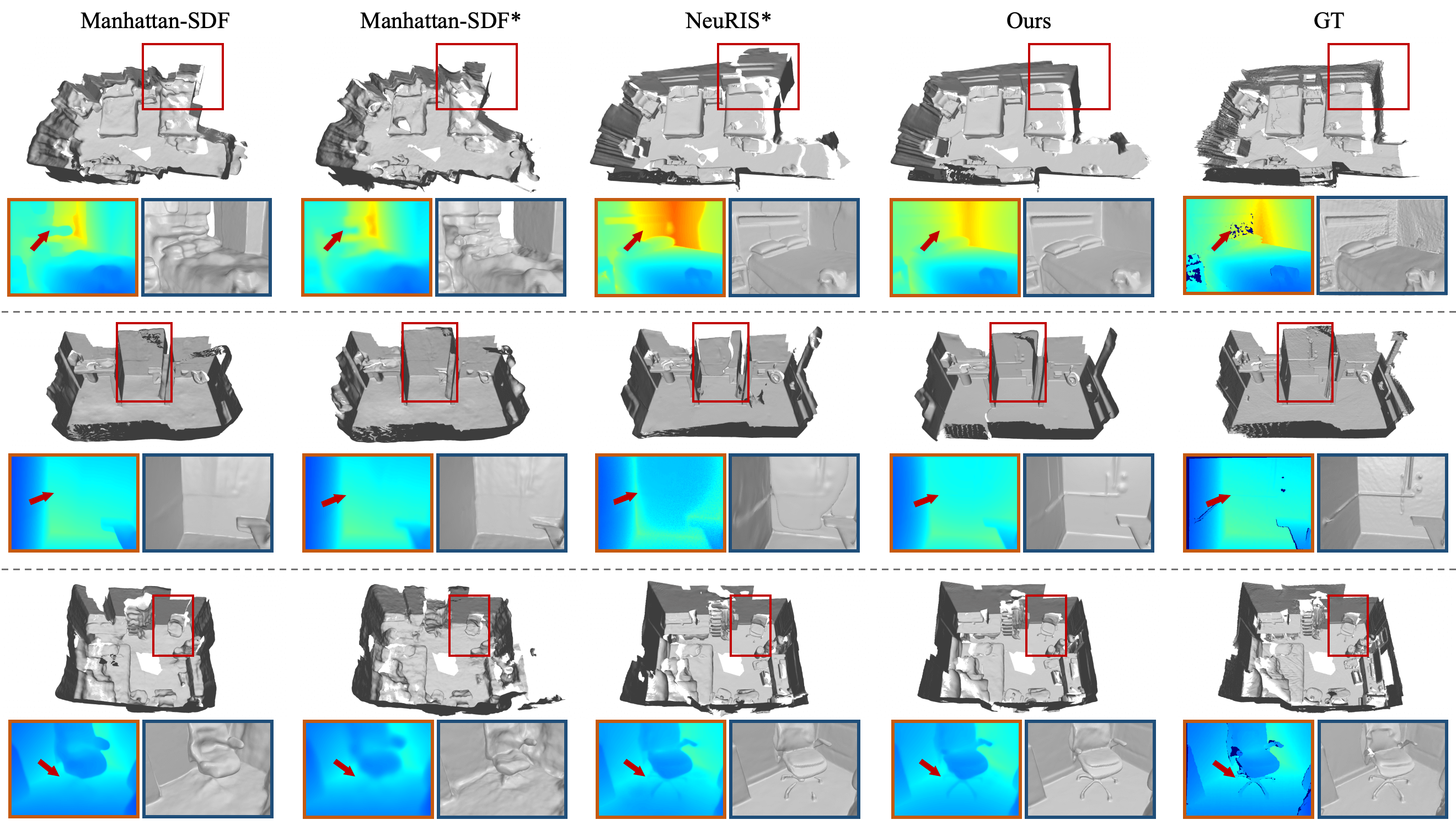}
    \vspace{-0.8cm}
    \caption{\textbf{Qualitative comparisons of surface reconstruction results.} We compare our surface reconstruction results to vanilla Manhattan-SDF \cite{Guo:etal:CVPR2022}, Manhattan-SDF*, and NeuRIS* (same geometry to NeuRIS \cite{wang2022neuris}). Both Manhattan-SDF and Manhattan-SDF* exhibit poor geometry and lose many object details (left two columns). Although estimated normals effectively aid surface reconstruction, NeuRIS* struggles on texture-less regions (third column). With the proposed SGR, our method produces superior results not only on texture-less regions (top two rows),  but also on object regions (last row).}
    \label{fig:Comparison:3DGeorecon}
    \vspace{-0.4cm}
\end{figure*}

\noindent \textbf{Baseline.} We compare our method to the following baselines: (1) \textbf{NeuRIS*}: As NeuRIS \cite{wang2022neuris} is only built for geometry reconstruction, we enhance NeuRIS with extra semantic attribute by integrating a semantic MLP. As discussed in Sec. \ref{section 3.1}, we also prevent gradients from the semantic loss (Eq. \ref{semantic loss}) back to SDF branch, and denote this version as NeuRIS*. (2)  \textbf{Manhattan-SDF} and \textbf{Manhattan-SDF*}: Since Manhattan-SDF \cite{Guo:etal:CVPR2022} mainly considers three coarse semantic classes: floor, wall and others, we extend it to learn NYU-40 classes and denote it as Manhattan-SDF* while keeping all other components intact. We also attempted to stop the gradient of semantic loss (Eq. \ref{semantic loss}) like NeuRIS* and ours, but found no significant gains for Manhattan-SDF*. (3) \textbf{S-NeRF}: We re-trained density-based Semantic-NeRF \cite{Zhi:etal:ICCV2021} using the same images and predicted labels.

%%%%%%%%%%%%%%%%%%%%%%%%%%%%%%%%%%%%%%%%%%%
\vspace{-0.1cm}
\subsection{Evaluation and Comparison}

\begin{table}[!t]
\caption{Quantitative Results of Fused 3D/2D Semantic Segmentation}
\vspace{-0.2cm}
\centering
    \resizebox{\linewidth}{!}{%
    \begin{tabular}{l|ccc|ccc}
    \toprule
    \multirow{2}{*}{Method} & \multicolumn{3}{c|}{3D Semantics} & \multicolumn{3}{c}{2D Semantics} \\ \cmidrule(l){2-7} 
     & Acc$\uparrow$ & mAcc$\uparrow$ & mIoU$\uparrow$ & Acc$\uparrow$ & mAcc$\uparrow$ & mIoU$\uparrow$ \\ \midrule
    DeeplabV3+ & \textbackslash{} & \textbackslash{} & \textbackslash{} & 0.613 & 0.763 & 0.503  \\
    S-NeRF & 0.471 & 0.644 & 0.361 & 0.635 & 0.783 & 0.532 \\
    Manhattan* & 0.628 & 0.672 & 0.466 & 0.643 & 0.793 & 0.545 \\
    NeuRIS* & \cellcolor{orange!25}0.631 & \cellcolor{orange!25}0.766 & \cellcolor{orange!25}0.529 & \cellcolor{orange!25}0.662 & \cellcolor{orange!25}0.819 & \cellcolor{orange!25}0.573 \\
    \textbf{Ours} & \cellcolor{red!25}\textbf{0.647} & \cellcolor{red!25}\textbf{0.789} & \cellcolor{red!25}\textbf{0.562} & \cellcolor{red!25}\textbf{0.693} & \cellcolor{red!25}\textbf{0.844} & \cellcolor{red!25}\textbf{0.619} \\ 
    \bottomrule
    \end{tabular}
    }\\
\label{Table:Comparison:SemSeg}
\vspace{-0.4cm}
\end{table}

\begin{table}[!t]
\caption{Quantitative Results of Surface Reconstruction}
\vspace{-0.2cm}
\centering
    \resizebox{\linewidth}{!}{%
    \begin{tabular}{l|ccccccc}
    \toprule
    Method & Acc$\downarrow$ & Comp$\downarrow$ & Prec$\uparrow$ & Recall$\uparrow$ & F-score$\uparrow$  & CD$\downarrow$ \\ 
    \midrule
    S-NeRF & 0.162 & 0.109 & 0.193 & 0.297 & 0.233 & 0.065 \\
    Manhattan & 0.065 & 0.066 & 0.642 & 0.615 & 0.628 & 0.025\\
    Manhattan* & 0.122 & 0.077 & 0.457 & 0.471 & 0.464 & 0.099\\
    NeuRIS* & \cellcolor{orange!25}0.051 & \cellcolor{orange!25}0.041 & \cellcolor{orange!25}0.746 & \cellcolor{orange!25}0.756 & \cellcolor{orange!25}0.751 & \cellcolor{orange!25}0.014\\
    \textbf{Ours} & \cellcolor{red!25}\textbf{0.045} & \cellcolor{red!25}\textbf{0.037} & \cellcolor{red!25}\textbf{0.773} & \cellcolor{red!25}\textbf{0.779} & \cellcolor{red!25}\textbf{0.776} & \cellcolor{red!25}\textbf{0.011}\\ 
    \bottomrule
    \end{tabular}
    }
\label{Table:Comparison:GeoRecon}
\vspace{-0.5cm}
\end{table}

\noindent \textbf{3D Semantic Segmentation.} Tab. \ref{Table:Comparison:SemSeg} presents the overall quantitative results of 3D semantic segmentation. Compared with baselines, our approach demonstrates significant improvement across all metrics. It is noteworthy that Manhattan-SDF* with extended 40 semantic classes yields relatively poor performance (46.6\% mIoU). This deficiency can be attributed to its joint optimization strategy, which only covers two planar semantic regions (i.e., wall, floor) and thus lacks constraints for finer-grained objects. Compared to Manhattan-SDF*, NeuRIS* exhibits higher semantic fusion performance (+6.3\% mIoU), benefiting from its more precise geometry through the utilization of normal priors. MOSE, built upon NeuRIS*, outputs a smoother 3D semantic map (further +3.3\% mIoU) thanks to proposed LCF and SGR modules. Qualitative results of 3D semantic reconstruction are presented in Fig. \ref{fig:Comparison:3DSemseg}. Inputting erroneous 2D labels with coarse neural priors, our MOSE reconstructs more accurate and consistent semantic maps, while Manhattan-SDF* and NeuRIS-S* struggle to ensure labels continuity.

\noindent \textbf{2D Semantic Segmentation.} 
We also render learned semantics into image space and report 2D semantic segmentation metrics in Table \ref{Table:Comparison:SemSeg}. Compared with the input DeeplabV3+ labels, both baseline methods and MOSE are capable of producing higher quality 2D semantic images, benefiting from the compact implicit scene representation. Among these methods, our MOSE achieves the best performance across all metrics, surpassing the second-highest NeuRIS* score by 4.6\% mIoU and outperforming the input DeeplabV3+ labels by 11.6\% mIoU. Fig. \ref{fig:Comparison:2DSemseg} presents qualitative results and ours are also visually much better than other methods by predicting coherent labels.

\begin{figure*}[!t]
    \centering
    \includegraphics[width=1\linewidth]{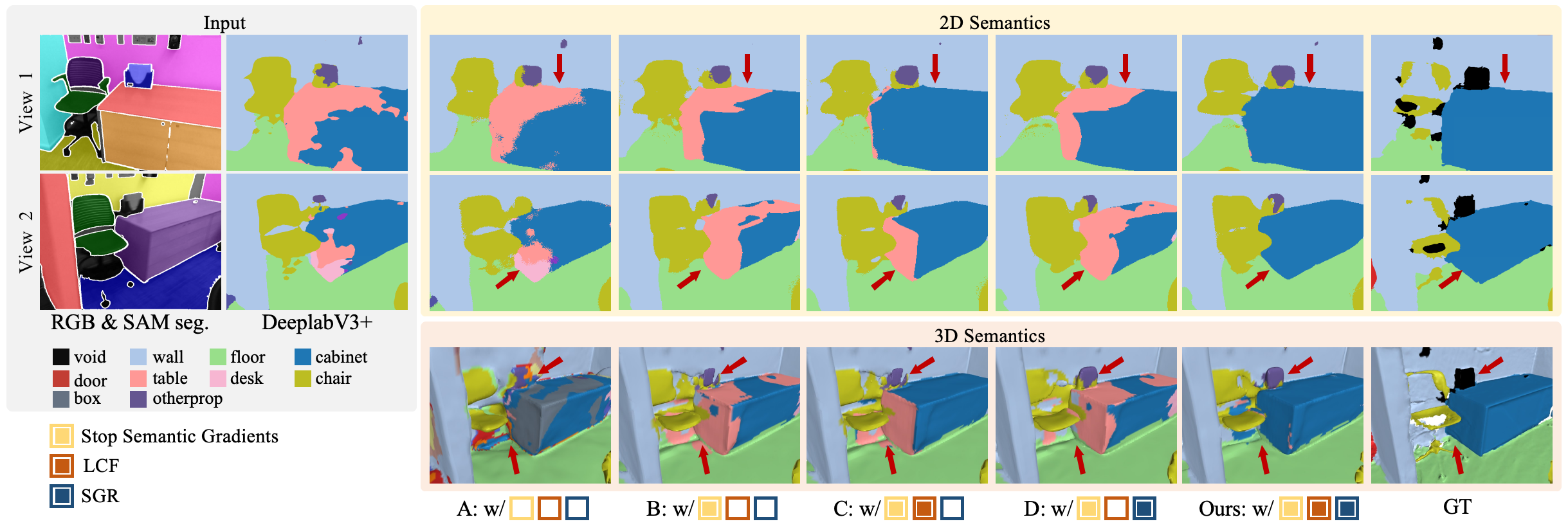}
    \vspace{-0.7cm}
    \caption{\textbf{Fused 3D/2D semantic results of ablation study.} Comparing Model-A with Model-B, the volume density (i.e., SDF value) often distorts
    to accommodate input noisy labels, and preventing SDF MLP from the influence of semantic loss brings significant benefits to the semantic field. Comparing Model-B with Model-C, after adopting LCF module, our method is able to produce more accurate and consistency semantic labels. Comparing Model-C with ours MOSE, with the help of SGR module, smoother semantic field can be learned benefiting from more precise radiance field. }
    \label{fig:Ablation:Semseg}
    \vspace{-0.4cm}
\end{figure*}

\noindent \textbf{3D Surface Reconstruction.} 
As demonstrated in Tab \ref{Table:Comparison:GeoRecon}, our method achieves the best reconstruction performance as well. Although Manhattan-SDF \cite{Guo:etal:CVPR2022} applies Manhattan constraints to regularize planar regions, and utilizes sparse depth from point cloud, both Manhattan-SDF and Manhattan-SDF* exhibit lower geometric quality compared to NeuRIS* and ours with normal priors. These results demonstrate the unique effectiveness of normal priors in multi-view 3D reconstruction, which has been validated by \cite{wang2022neuris, park:ICLR2024:h2oSDF}. \cite{Guo:etal:CVPR2022} lacks supervision for objects where Manhattan assumption is not satisfied (first row of Fig. \ref{fig:Comparison:3DGeorecon}). Please note that NeuRIS* shares the same geometric quality to NeuRIS, as the SDF MLP receives no gradients from semantics. MOSE further enhances the geometric quality (+2.5\% F-score compared to NeuRIS*). As shown in Fig. \ref{fig:Comparison:3DGeorecon}, MOSE shows better visualization results not only on texture-less regions (top two rows) but also on more object details (last row).

\vspace{-0.2cm}
\subsection{Ablation Studies}

\begin{figure}[!t]
    \centering\includegraphics[width=1.00\linewidth]{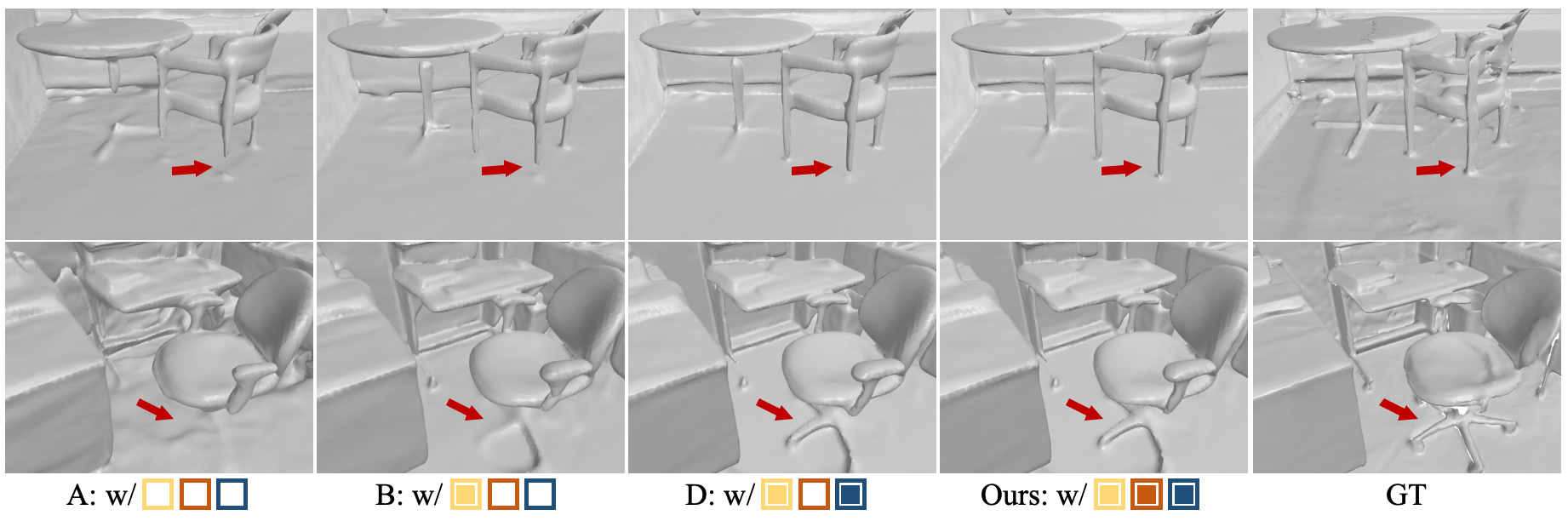}
    \vspace{-0.6cm}
    \caption{\textbf{Surface reconstruction results of ablation study}. Comparing Model-A with Model-B, preventing semantic gradients to other branches leads to significant improvements in surface reconstruction. Compared with Model-B, ours preserve more object details with the help of SGR module. Model-C is not presented here as LCF module does not affect SDF field.}
    \label{fig:Ablation:Georecon}
    \vspace{-0.2cm}
\end{figure}

\begin{table}[!t]
\caption{Ablation Studies of Our Design Choices}
\vspace{-0.1cm}
\centering
    \resizebox{\linewidth}{!}{%
    \begin{tabular}{cccc|cccc}
    \toprule
    Model & Stop & LCF & SGR & mIoU(3D)$\uparrow$ & mIoU(2D)$\uparrow$ & F-Score$\uparrow$ & CD$\downarrow$ \\ 
    \midrule
    A & \XSolidBrush& \XSolidBrush & \XSolidBrush & 0.328 & 0.532 & 0.617 & 0.023 \\
    B & \Checkmark & \XSolidBrush & \XSolidBrush & 0.529 & 0.573 & 0.751 & \cellcolor{orange!25}0.014 \\
    C & \Checkmark & \Checkmark & \XSolidBrush & \cellcolor{orange!25}0.553 & \cellcolor{orange!25}0.597 & 0.751 & \cellcolor{orange!25}0.014 \\
    D & \Checkmark & \XSolidBrush & \Checkmark & 0.529 & 0.580 & \cellcolor{orange!25}0.775 & \cellcolor{red!25}\textbf{0.011} \\
    \textbf{Ours} & \Checkmark & \Checkmark & \Checkmark & \cellcolor{red!25}\textbf{0.562} & \cellcolor{red!25}\textbf{0.619} & \cellcolor{red!25}\textbf{0.776} & \cellcolor{red!25}\textbf{0.011} \\ \bottomrule
    \end{tabular}
    }
\label{Table:Ablation}
\vspace{-0.4cm}
\end{table}

In order to evaluate the effectiveness of each component in our method, we conduct ablation studies with different settings and variants: \textbf{Model-A}: directly integrate a semantic MLP into NeuRIS; \textbf{Model-B}: prevent SDF MLP of Model-A from the influence of semantics (i.e, NeuRIS*). Thus Model-B  has the same geometry quality to NeuRIS \cite{wang2022neuris}; \textbf{Model-C}: built upon Model-B, with the LCF module. Note that we use Eikonal loss (Eq. \ref{eikonal loss}) to regularize the SDF field in Model-A, Model-B and Model-C; \textbf{Model-D}: built upon Model-B, with the SGR module; \textbf{Ours}: integrate both LCF and SGR modules. We report quantitative results in Tab \ref{Table:Ablation} and present qualitative results in Fig. \ref{fig:Ablation:Semseg} and Fig. \ref{fig:Ablation:Georecon}.

\noindent \textbf{Stopping Semantic Gradient.} Comparing Model-A against Model-B, we observe that preventing semantic gradients to other branches leads to improvements in both geometry and semantics evaluation. We attribute this to the negative impact of altering geometry to over-fit erroneous 2D labels, hence adhering to the radiance field of NeuRIS is more accurate.

\noindent \textbf{Locally-Consistency Fusion Strategy.} As LCF module concentrates on semantics, here we focus on the tasks of 3D and 2D semantic segmentation. Comparing Model-C with Model-B, the LCF module brings significant improvement in both 3D and 2D semantics. Fig. \ref{fig:Ablation:Semseg} also shows that inconsistent input 2D semantics under different views lead to a blending of two labels for the same object, while our proposed LCF is able to mitigate this effect and produce accurate labels.

\noindent \textbf{Semantically-Weighted Geometric Regularization.} Comparing Model-C with ours in Tab. \ref{Table:Ablation}, our designed SGR module brings benefits to both geometry and semantics. We observe that Model-D achieves almost the same geometry evaluation metrics as ours. We attribute this to the fact that SGR focuses on planar and object regions, and the fused semantics are already sufficiently accurate even without LCF module. Fig. \ref{fig:Comparison:3DGeorecon} (Note that Model-B is equivalent to NeuRIS*) and Fig. \ref{fig:Ablation:Georecon} shows that our SGR module not only effectively improves the reconstruction quality of planar regions, but also preserves more object details. In addition, benefiting from more accurate radiance field, MOSE is able to produce smoother semantics, especially when combined with LCF module (second-to-last columns of Fig. \ref{fig:Ablation:Semseg}).

\vspace{-0.1cm}
\subsection{Further Discussion} \label{discussion}

\begin{figure}[!t]
    \centering\includegraphics[width=1.00\linewidth]
    {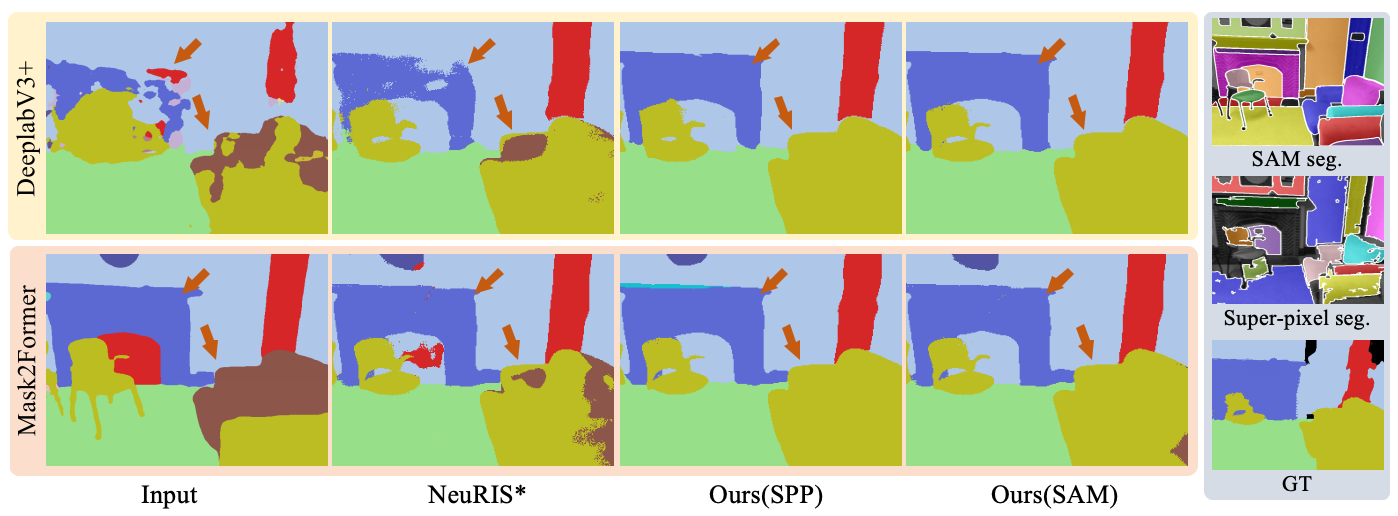}
    \vspace{-0.6cm}
    \caption{\textbf{Qualitative results of different labels and segments}. We validate the effectiveness of MOSE by inputting Mask2Former \cite{cheng2022mask2former} labels and super-pixel segments \cite{felzenszwalb2004efficient}. Qualitative results demonstrate that our method is not sensitive to prior predictors.}
    \label{fig:validation}
    \vspace{-0.2cm}
\end{figure}

We also discuss the sensitivity of our system to different input labels and segmentation priors using labels and segments from Mask2Former \cite{cheng2022mask2former} and unsupervised super-pixel techniques \cite{felzenszwalb2004efficient}. We use the COCO pre-trained Mask2Former \cite{lin2014coco,siddiqui:CVPR2023:panopticlifting} and remap its output semantic labels to NYU-40 classes. Therefore, its predictions under NYU-40 convention are relatively lower than fine-tuned DeepLabV3+. For 2D segment masks, we take graph-based super-pixels and remove those smaller than 4000 pixels. We focus on evaluating their impact on semantics as the geometry quality is almost comparable. Qualitative results in Fig. \ref{fig:validation} show that MOSE still exhibits commendable semantic fusion capability, indicating its robustness to input label qualities. Using super-pixels results in slightly inferior scores compared to those obtained using SAM, mainly due to less accurate segment masks. Tab. \ref{table:validation} confirms our observations and shows that our approach is not sensitive to the specific choices of prior predictors and maintains consistent performance across various types of prior predictors.

%%%%%%%%%%%%%%%%%%%%%%%%%%%%%%%%%%%%%%%%%%%%%%%%%%%%%%%%%%%%%%%%%%%%%%%%%%%%%%%%
\section{LIMITATIONS AND CONCLUSION}
In this work, we proposed MOSE, a NeRF-based 3D scene understanding approach only using multi-view images and 2D priors. To address the particularly limited performance of existing works on object-level semantics as well as geometry of texture-less regions, we introduce segment-guided consistency and semantic-guided smoothness to improve these capability. Both quantitative and qualitative results on the real-world ScanNet dataset have validated the effectiveness of MOSE, achieving promising results in both semantics and geometry. 

\noindent \textbf{Limitations.} Although MOSE demonstrates robustness to noisy priors, its performance may degrade when neural priors are incorrectly predicted across most viewpoints. Similar to other NeRF-based methods, MOSE still requires provided camera poses and relatively lengthy per-scene optimization, while in practice it would be more desirable if multi-view information could be efficiently accumulated to form a 3D model. Therefore, choosing the recently popular 3D Gaussian Splatting \cite{kerbl:2023:3dgs} as our novel backbone or conducting joint optimization of MLPs and camera states could serve as our valuable future work.

\begin{table}[!t]
\caption{Quantitative Results of Different labels And Segments}
\vspace{-0.2cm}
\centering
    \resizebox{\linewidth}{!}{%
    \begin{tabular}{l|ccc|ccc}
    \toprule
    \multirow{2}{*}{Method} & \multicolumn{3}{c|}{DeeplabV3+} & \multicolumn{3}{c}{Mask2Former} \\ \cmidrule(l){2-7} 
     & mIoU(3D)$\uparrow$ & mIoU(2D)$\uparrow$ & F-score$\uparrow$ & mIoU(3D)$\uparrow$ & mIoU(2D)$\uparrow$ & F-score$\uparrow$ \\ \midrule
    Input & \textbackslash{} & 0.503 & \textbackslash{} &\textbackslash{} & 0.440 & \textbackslash{} \\
    NeuRIS* & 0.529 & 0.573 & 0.751 & 0.427 & 0.471 & 0.751 \\
    Ours(SPP) & \cellcolor{orange!25}0.539 & \cellcolor{orange!25}0.596 & \cellcolor{red!25}\textbf{0.777} & \cellcolor{orange!25}0.434 & \cellcolor{orange!25}0.478 & \cellcolor{red!25}\textbf{0.777} \\
    Ours(SAM) & \cellcolor{red!25}\textbf{0.562} & \cellcolor{red!25}\textbf{0.619} & \cellcolor{orange!25}0.776 & \cellcolor{red!25}\textbf{0.450} & \cellcolor{red!25}\textbf{0.489} & \cellcolor{orange!25}0.772 \\ \bottomrule
    \end{tabular}
    } \\
    \label{table:validation}
    \vspace{-0.4cm}
\end{table}

%                                   % the textheight of the last page by a suitable amount.
%                                   % This command does not take effect until the next page
%                                   % so it should come on the page before the last. Make
%                                   % sure that you do not shorten the textheight too much.

%%%%%%%%%%%%%%%%%%%%%%%%%%%%%%%%%%%%%%%%%%%%%%%%%%%%%%%%%%%%%%%%%%%%%%%%%%%%%%%%
% \section*{ACKNOWLEDGMENT}
%%%%%%%%%%%%%%%%%%%%%%%%%%%%%%%%%%%%%%%%%%%%%%%%%%%%%%%%%%%%%%%%%%%%%%%%%%%%%%%%
\bibliographystyle{IEEEtran}
\bibliography{main}
%%%%%%%%%%%%%%%%%%%%%%%%%%%%%%%%%%%%%%%%%%%%%%%%%%%%%%%%%%%%%%%%%%%%%%%%%%%%%%%%
\end{document}